\documentclass[10pt,twocolumn,letterpaper]{article}

\usepackage[pagenumbers]{cvpr} 

\usepackage[dvipsnames]{xcolor}

\definecolor{cvprblue}{rgb}{0.21,0.49,0.74}
\usepackage[pagebackref,breaklinks,colorlinks,citecolor=cvprblue]{hyperref}
\usepackage{algorithmic}
\usepackage{algorithm}

\usepackage{listings}

\def\confName{CVPR}
\def\confYear{2024}
\usepackage[capitalize]{cleveref}
\usepackage{tikz}
\usepackage{xcolor}
\usepackage{multirow}
\usepackage{xcolor,colortbl}
\usepackage{color}
\usepackage[accsupp]{axessibility}
\definecolor{codegreen}{rgb}{0,0.5,0.5}
\definecolor{codegray}{rgb}{0.5,0.5,0.5}
\definecolor{codepurple}{rgb}{0.58,0,0.82}
\definecolor{backcolour}{rgb}{1, 1, 1}

\lstdefinestyle{mystyle}{
    backgroundcolor=\color{backcolour},   
    commentstyle=\color{codegreen},
    keywordstyle=\color{magenta},
    numberstyle=\tiny\color{codegray},
    stringstyle=\color{codepurple},
    basicstyle=\ttfamily\footnotesize,
    breakatwhitespace=false,         
    breaklines=true,                 
    captionpos=b,                    
    keepspaces=true,                 
    numbers=none,                    
    numbersep=5pt,                  
    showspaces=false,                
    showstringspaces=false,
    showtabs=false,                  
    tabsize=2
}
\newcommand*\circled[1]{\tikz[baseline=(char.base)]{
            \node[shape=circle,draw,inner sep=0.3pt] (char) {#1};}}

\title{
UniMix: Towards Domain Adaptive and Generalizable LiDAR Semantic Segmentation in Adverse Weather
}

\author{Haimei Zhao$^{1\star}$\quad
 Jing Zhang$^{1\star}$ \quad Zhuo Chen$^{2}$\quad Shanshan Zhao$^{1}$ \quad Dacheng Tao$^{3}$\\
  $^{1}$The University of Sydney \quad $^{2}$
  Tsinghua University \quad $^{3}$ Nanyang Technological University\\
  {\tt\small \{hzha7798, jing.zhang1\}@sydney.edu.au, z-chen17@mails.tsinghua.edu.cn,}\\
  {\tt\small \{sshan.zhao00, dacheng.tao\}@gmail.com}
}

\begin{document}
\maketitle
\def\thefootnote{$\star$}
\footnotetext{These authors contributed equally to this work.}
\def\thefootnote{\arabic{footnote}}
\begin{abstract}
LiDAR semantic segmentation (LSS) is a critical task in autonomous driving and has achieved promising progress. However, prior LSS methods are conventionally investigated and evaluated on datasets within the same domain in clear weather. The robustness of LSS models in unseen scenes and all weather conditions is crucial for ensuring safety and reliability in real applications. To this end, we propose UniMix, a universal method that enhances the adaptability and generalizability of LSS models. UniMix first leverages physically valid adverse weather simulation to construct a Bridge Domain, which serves to bridge the domain gap between the clear weather scenes and the adverse weather scenes. Then, a Universal Mixing operator is defined regarding spatial, intensity, and semantic distributions to create the intermediate domain with mixed samples from given domains. Integrating the proposed two techniques into a teacher-student framework, UniMix efficiently mitigates the domain gap and enables LSS models to learn weather-robust and domain-invariant representations. We devote UniMix to two main setups: 1) unsupervised domain adaption, adapting the model from the clear weather source domain to the adverse weather target domain; 2) domain generalization, learning a model that generalizes well to unseen scenes in adverse weather. Extensive experiments validate the effectiveness of UniMix across different tasks and datasets, all achieving superior performance over state-of-the-art methods. The code will be released.
\end{abstract}    
\section{Introduction}
\label{sec:intro}
LiDAR semantic segmentation (LSS) is a fundamental task of 3D scene understanding for autonomous driving, aiming to assign a semantic label for each 3D point.  The progress in deep learning techniques, coupled with the availability of large-scale datasets, has propelled LSS methods to deliver promising results. However, traditional LSS models are typically trained and evaluated using data collected in clear weather conditions, assuming a consistent domain and lacking domain adaptive and generalizable ability. This limitation poses challenges in real-world applications where autonomous driving systems encounter diverse scenes and weather conditions, each characterized by distinct data distributions. Adverse weather, in particular, introduces variations in the spatial positions, intensity values, and semantic distributions of LiDAR point clouds \cite{dreissig2023survey, montalban2021quantitative}.  Models trained in ideal conditions often perform inadequately in adverse weather scenarios \cite{xiao20233d}. Therefore, the robust handling of unseen scenes in diverse weather is essential for ensuring the reliability and safety of autonomous driving.

\begin{figure}[t]
    \begin{center}
    \includegraphics[width=0.8\linewidth]{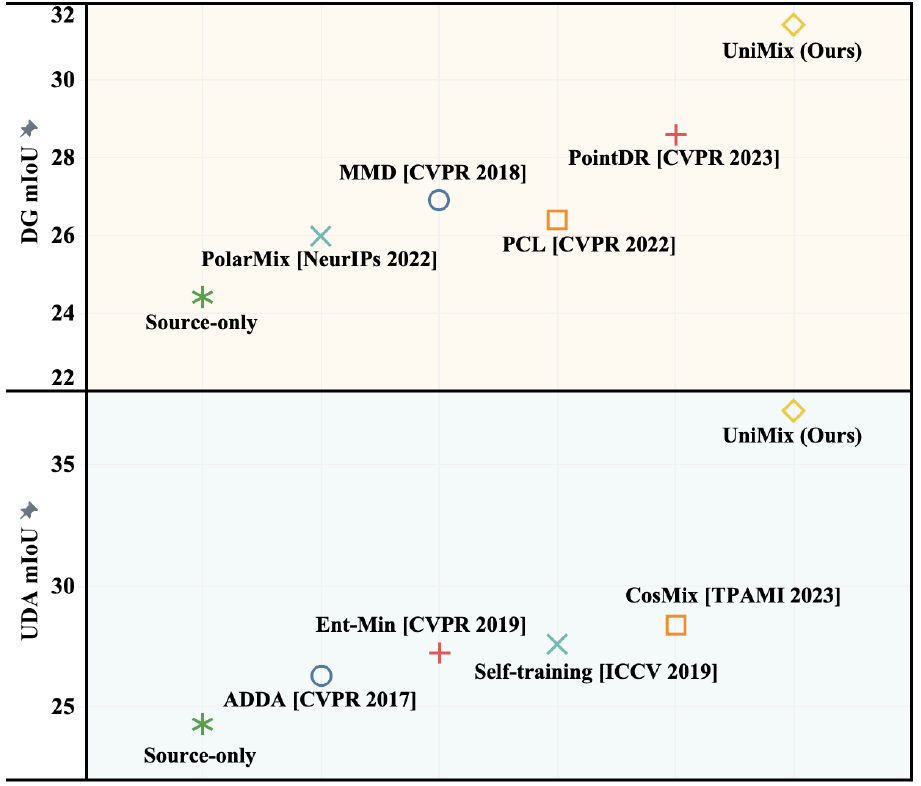}
    \end{center}
    \vspace{-0.45cm}
    \caption{UniMix outperforms SOTA methods in both UDA and DG tasks, using SemanticKITTI~\cite{behley2019semantickitti} as the source and SemanticSTF~\cite{xiao20233d} in all four weather conditions as the target.}
    \label{figure:startfig}
    \vspace{-0.3cm}
\end{figure}

While domain adaptation and generalization techniques have proven effective in 2D semantic segmentation \cite{Ma_2022_CVPR,bruggemann2023refign,gong2023train}, their application in LSS has primarily focused on adapting or generalizing between synthetic and real scenes or across different real-world scenes~\cite{saltori2022cosmix_eccv,saltori2022cosmix}. This leaves a gap in the exploration for handling unseen scenes in adverse weather. To be specific, the domain discrepancy arises from significant semantic distribution variations between datasets captured in different scenes, such as structural layout differences in various cities. Furthermore, adverse weather introduces spatial noise and impacts point cloud reflection intensity, resulting in shifts in data distribution—another form of domain discrepancy. The coexistence of these dual sources of domain discrepancy presents formidable challenges for adapting and generalizing existing LSS models.
To address these challenges, we introduce UniMix, a universal approach for learning weather-robust and domain-invariant representations, enabling the adaptation and generalization of LSS models from clear to adverse weather scenes. UniMix is structured as a two-stage framework, consisting of Source-to-Bridge and Bridge-to-Target stages, grounded in the well-established teacher-student framework~\cite{saltori2022cosmix_eccv,saltori2022cosmix,kong2023lasermix}. First, we construct a Bridge Domain through physically realistic weather simulation on source domain data, generating data with the same scene characteristics but in diverse adverse weather conditions. We then introduce a Universal Mixing operator that blends point clouds from two given domains through spatial, intensity, and semantic mixing. These two techniques significantly mitigate the domain gap and enhance scene diversity during model training, thereby improving the model's adaptability and generalizability. Specifically, in the first Source-to-Bridge stage, even without access to target data, the model learns weather-robust representations from the mixed source and Bridge Domain data via Universal Mixing, enabling effective generalization to unseen adverse weather conditions. In the second Bridge-to-Target stage, with access to target data, the model further adapts to the target domain and learns domain-invariant representations from the mixed Bridge Domain and target domain data. UniMix thus offers a versatile solution and can be devoted to two tasks: 1) unsupervised domain adaptation (UDA) and 2) domain generalization (DG), boosting the performance of LSS models across diverse weather conditions and scenes.

We conduct a thorough evaluation of UniMix on both UDA and DG tasks. We employ large-scale public LiDAR segmentation benchmarks, utilizing SemanticKITTI \cite{behley2019semantickitti} and SynLiDAR \cite{xiao2022transfer} as the source datasets and SemanticSTF \cite{xiao20233d} as the target dataset. As shown in Fig. \ref{figure:startfig}, experimental results for both tasks demonstrate UniMix's efficacy in mitigating the complex domain gap arising from diverse scenes and adverse weather conditions, surpassing the performance of state-of-the-art (SOTA) methods. Furthermore, we present comprehensive analysis and ablation studies on its key components to validate their effectiveness.

In summary, our contributions are as follows:
\begin{itemize}
\item We propose UniMix, a universal method that enhances the adaptability and generalizability of LSS models to unseen adverse-weather scenes.

\item We construct a Bridge Domain through physically realistic weather simulation, bridging the gap between the clear weather scenes and the adverse weather scenes.

\item We present Universal Mixing, which generates diverse intermediate point clouds from given domains to mitigate domain gaps caused by different scenes and weather.

\item UniMix shows its effectiveness and achieves state-of-the-art performance in both UDA and DG tasks.
\end{itemize}


\section{Related Work}
\label{sec:relatedwork}

\textbf{LiDAR Semantic Segmentation} refers to predicting semantic labels for each 3D point in LiDAR point clouds. It has been extensively investigated in existing approaches with various network architectures and from different perspectives. Approaches can be categorized into point-based (directly handling irregular point cloud data using MLPs \cite{qi2017pointnet,qi2017pointnet++}, convolution networks \cite{li2018pointcnn,su2018splatnet,liu2019point,thomas2019kpconv}, and transformers \cite{guo2021pct,duan2024condaformer}), 2D projection-based (saving computation costs by projecting points to range-view images \cite{wu2018squeezeseg,wu2019squeezesegv2,milioto2019rangenet++,cortinhal2020salsanext} or bird's-eye-view \cite{zhang2020polarnet}), and voxel-based (representing point clouds as 3D voxels for employing dense 3D convolutions \cite{wu20153d,qi2016volumetric} or sparse convolutions \cite{graham20183d,choy20194d}).

\textbf{Unsupervised Domain Adaptation} aims to transfer knowledge to address domain shift between a labeled source domain and an unlabeled target domain. In the field of LSS, some approaches \cite{wu2019squeezesegv2,zhao2021epointda,rochan2022unsupervised,kong2023conda} project point clouds to 2D images, leveraging established UDA techniques for 2D semantic segmentation. Alternatively, some methods perform direct domain transfer in 3D space through canonical domain construction \cite{Yi_2021_CVPR}, domain mapping \cite{langer2020domain,xiao2022transfer}, or domain mixing \cite{saltori2022cosmix}. Notably, CoSMix \cite{saltori2022cosmix} utilizes point cloud mixing by assembling semantic patches from different domains, demonstrating the efficacy of domain mixing. While these UDA methods make strides in real-to-real and synthetic-to-real adaptation tasks, none specifically addresses clear-to-adverse weather adaptation, which is crucial for mitigating adverse weather effects in real-world applications. Our method is tailored for domain adaptation and generalization to unseen scenes in adverse weather.

\textbf{Domain Generalization} seeks to train a model using only source domain data that can generalize well to unseen target domains. In the 2D vision domain, it has been extensively investigated in previous research \cite{balaji2018metareg,li2018domain,zhao2020domain,choi2021robustnet,zhou2022domain}. To address domain shift in 3D scene understanding within LSS, some DG methods have been proposed recently. Kim et al. \cite{kim2023single} present a single-domain generalization method that enforces consistency between the source and augmented domain, generalizing from a source domain to a target domain with distinct LiDAR configuration and scenes. LiDOG \cite{saltori2023walking} augments the learning network to be robust to sensor placement shifts and resolutions by incorporating an auxiliary BEV segmentation branch. While these methods prove effective in synthetic-to-real and real-to-real generalization, they focus on clear weather conditions. Xiao et al. \cite{xiao20233d} explore adverse weather generalization by creating additional augmented point cloud views to train the model for learning perturbation-invariant representations. Our method also targets learning weather-robust and domain-invariant representations and universally addresses domain adaptation and generalization to adverse weather scenes.


\textbf{3D Perception in Adverse Conditions} has gained significant attention among both academic and industrial researchers \cite{dreissig2023survey,dong2023benchmarking,zhu2023understanding} due to its practical significance. Synthetic data exploration in adverse weather conditions, such as fog and snowfall, has been conducted by Kilic et al. \cite{kilic2021lidar} and Hahner et al. \cite{hahner2021fog,hahner2022lidar}, who propose robust 3D object detection methods. For real data investigations, Bijelic et al. \cite{bijelic2020seeing} introduce the STF dataset, a large multimodal dataset for 3D object detection in adverse weather, along with a fusion-based model. SemanticSTF \cite{xiao20233d} extends STF \cite{bijelic2020seeing} with additional semantic labels, facilitating the study of LiDAR semantic segmentation in adverse conditions. In this study, we adopt SemanticSTF as the target domain for the domain adaptation and generalization experiments.


\section{Method}

\begin{figure}[t]
    \begin{center}
    \includegraphics[width=\linewidth]{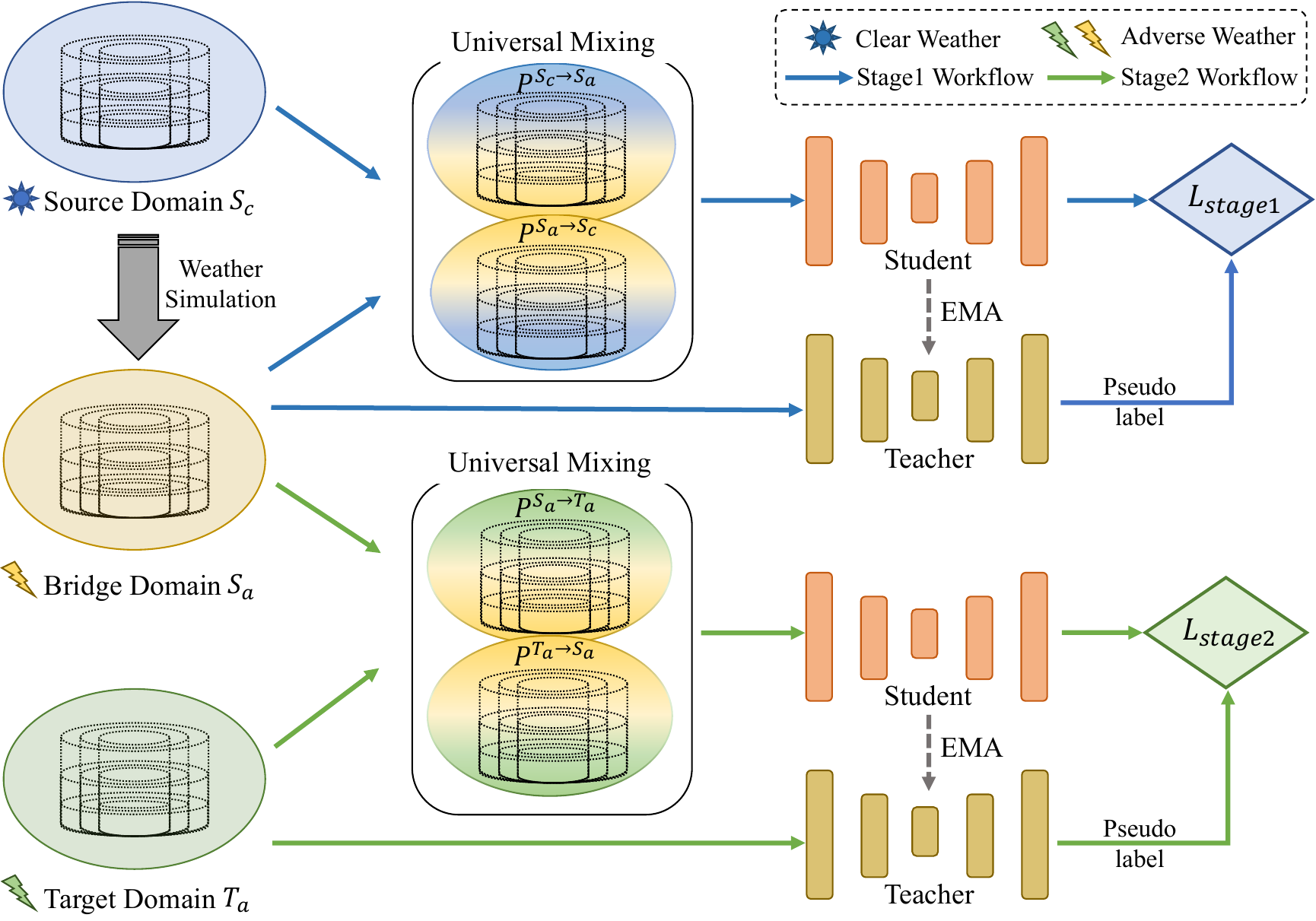}
    \end{center}
    \vspace{-0.45cm}
    \caption{The overall pipeline of UniMix. In the first stage (top part), the clear-weather source domain $S_c$ and the simulated adverse-weather Bridge Domain $B_a$ are taken as input to generate the intermediate domain $\{P^{S_c \rightarrow B_a}, P^{B_a \rightarrow S_c}\}$ via Universal Mixing, and the student network is trained under the supervision of $L_{stage1}$. In the second stage (bottom part), the Bridge Domain and adverse-weather target domain $T_a$ are utilized to generate the intermediate domain $\{P^{B_a \rightarrow T_a}, P^{T_a \rightarrow B_a}\}$ via Universal Mixing, and the student network is trained under the supervision of $L_{stage2}$. The teacher is leveraged to produce pseudo labels and is updated via EMA \cite{tarvainen2017mean} of the student's weights.}
    \label{figure:mainfigure}
    \vspace{-0.3cm}
\end{figure}

UniMix offers a universal approach to robust 3D representation learning for LiDAR semantic segmentation. It addresses domain discrepancies between clear and adverse weather conditions, serving both domain adaptation and generalization tasks. Specifically, two types of domain gaps between source and target domains are considered: scene-level and weather-level. To this end, we first introduce a Bridge Domain through physically valid weather condition simulation on source data (Sec.~\ref{secBridge}). Then, we present a Universal Mixing operator to comprehensively blend points across domains, enriching point cloud data in different spatial, intensity, and semantic distributions (Sec.~\ref{secUMO}). Utilizing a teacher-student framework~\cite{saltori2022cosmix_eccv}, UniMix proves effective for both UDA and DG tasks. The overall pipeline is outlined in Fig. \ref{figure:mainfigure} and Algo. \ref{alg:unimix} (Sec.~\ref{secNets}).

\subsection{Bridge Domain Generation} \label{secBridge}

\noindent\textbf{Effect of Adverse Weather.}
Previous studies have shown that adverse weather significantly impacts active pulsed systems like LiDAR and hinders the performance of traditional perception models~\cite{dong2023benchmarking,zhu2023understanding}. In adverse weather such as fog, rain, and snow, airborne particles interact with the laser beam, absorbing, reflecting, or refracting its photons \cite{dreissig2023survey,hahner2021fog,hahner2022lidar}. This interaction notably affects the intensity values, point count, spatial allocations, and semantic distributions of point clouds. To address this, we propose creating a Bridge Domain by weather simulation to help train adaptive and generalizable LSS models for adverse weather.

\noindent\textbf{Bridge Domain.}
According to prior studies \cite{rasshofer2011influences,hahner2021fog,hahner2022lidar}, the LiDAR system can be represented using a linear model for the received signal power. For non-elastic scattering, the range-dependent received power $P_R$ is modeled as the time-wise convolution of the time-dependent transmitted signal power $P_T$ and the optical system's impulse response $H$:
\begin{equation} \label{eq:LiDARequation}
    P_R(R) = C_A \int_0^{2R/c} P_T(t)\,H\left(R-\frac{ct}{2}\right) dt,
\end{equation} 
where $R$ is the object distance, $c$ is the speed of light and $C_A$ is a system constant independent of time $t$ and range. 

\begin{figure}[t]
    \begin{center}
    \includegraphics[width=\linewidth]{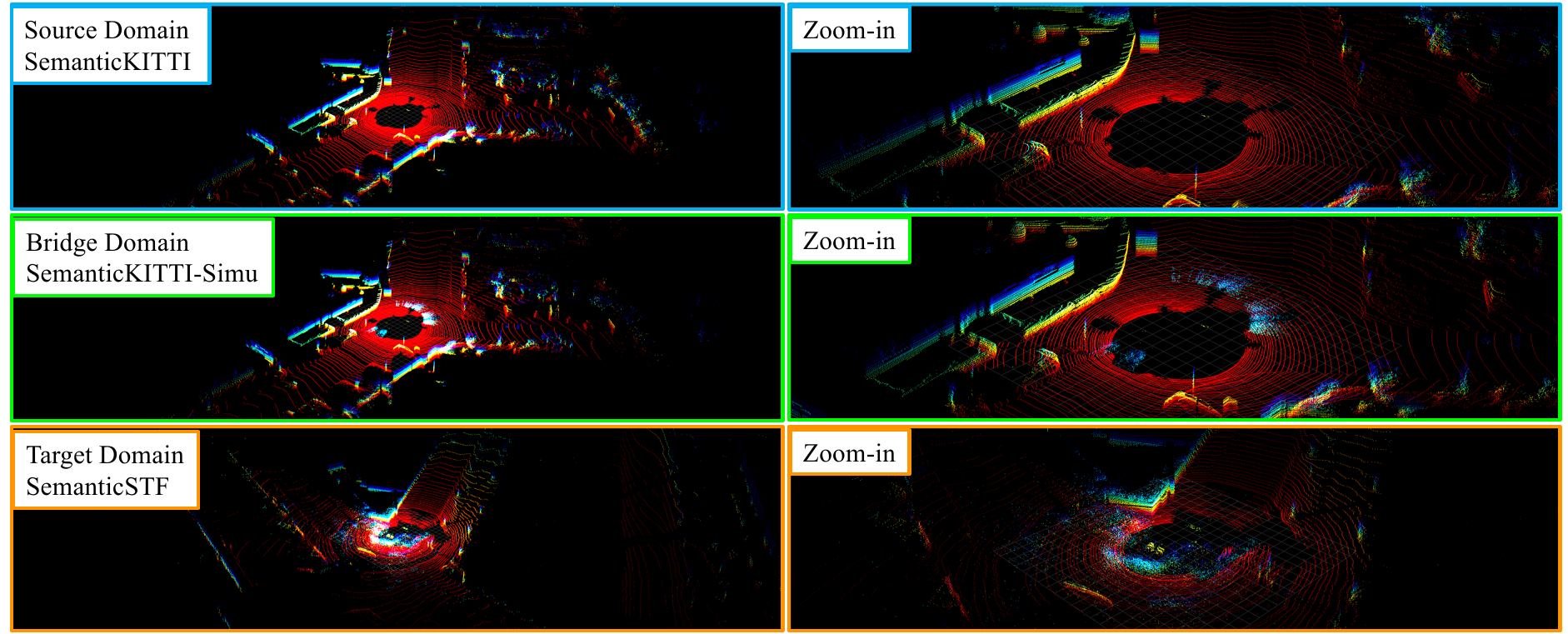}
    \end{center}
    \vspace{-0.5cm}
    \caption{Visualization of point clouds in the source domain, Bridge Domain, and target domain. The Bridge and target domain are in light fog weather. Better viewed with zoom-in and in color.}
    \label{figure:simueffect}
    \vspace{-0.3cm}
\end{figure}

In challenging weather, adverse conditions affect laser pulse $H$ due to scattering caused by rain, fog, and snowflakes, each with distinct properties~\cite{gunn1958distribution, rasshofer2011influences}, \eg, rain and fog consist of small water droplets and snowflakes contain ice crystals with varying densities. These scatterers, modeled as a statistical distribution of reflective spherical objects with varying diameters, attenuate the original power reflectance from solid objects. In this study, we adopt physically valid weather simulation methods \cite{hahner2021fog, hahner2022lidar} to simulate scatterers in the air and the wet ground effect. Specifically, we consider four frequent weather conditions in real life — dense fog, light fog, snow, and rain. Simulating these conditions for each scene in the source domain constructs the Bridge Domain, incorporating both the scene characteristics of the source domain and the weather features of the target domain. This process can be formulated as:
\begin{equation}
    B_a =(P^{B_a} , Y^{B_a} ) = \Omega(P^{S_c} , Y^{S_c} ).
\end{equation} 
Here, we use $\Omega$ to denote the overall weather simulation. The subscripts $c$ and $a$ denote clear weather and adverse weather, respectively. $S_c$ denotes the source domain while $B_a$ denotes the generated Bridge Domain. $P$ and $Y$ denote the point clouds and corresponding labels, respectively. We visualize the comparison of point clouds in the source domain, Bridge Domain, and target domain in Fig.~\ref{figure:simueffect}. The Bridge Domain keeps the scene characteristic of the source domain while integrating the weather disturbance (fog particles in blue) encountered in the target domain. More weather simulation effects can be found in the Appendix.


\subsection{Universal Mixing} \label{secUMO}
Since point cloud mixing techniques have been proven effective in various LSS settings before, including CoSMix (synthetic-to-real adaptation), PolarMix (data augmentation), and LaserMix (Semi-supervised LSS), we further dig into point cloud sample mixing according to the impact from diverse weather conditions. In our specific clear-to-adverse scenario, we conduct point cloud mixing regarding spatial, intensity, and semantic distributions to enrich the diversity of mixed intermediate point clouds and help models learn weather-robust and domain-invariant representations.

\begin{figure}[t]
    \begin{center}
    \includegraphics[width=\linewidth]{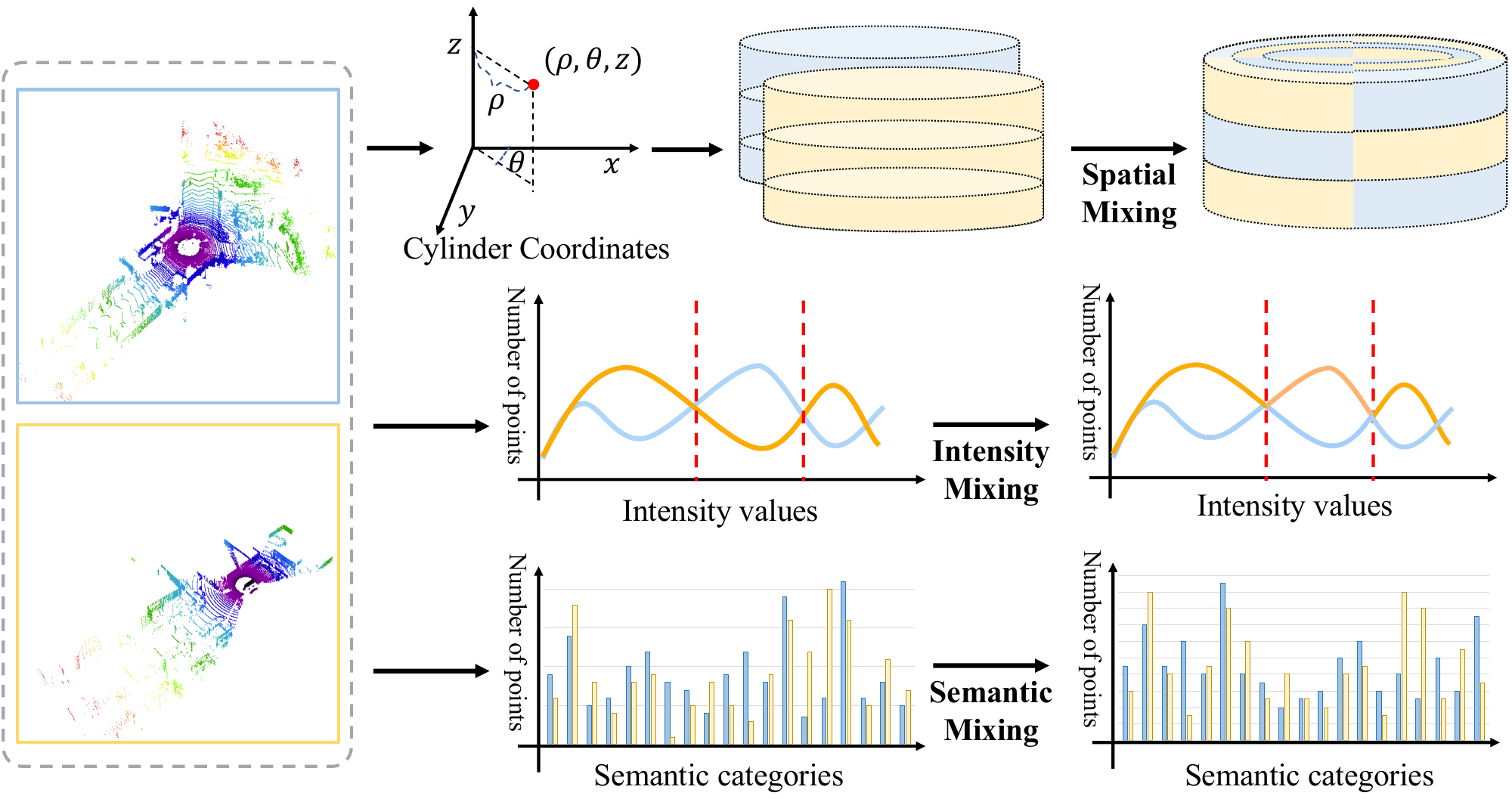}
    \end{center}
    \vspace{-0.5cm}
    \caption{Illustration of different sample mixing methods, including spatial mixing, intensity mixing, and semantic mixing.}
    \label{fig:mixingfig}
    \vspace{-0.3cm}
\end{figure}

Let $P \in \mathbb{R}^{N \times 4}$ denote a LiDAR point cloud with $N$ points whose label is represented by $Y$. Each point $p_i$ can be expressed as $(x_i,y_i,z_i, I_i)$, where $(x_i,y_i,z_i)$ is the 3D Cartesian coordinate of the point and $I_i$ is its intensity value received from the laser scanner. Given a source domain $S=(P^S, Y^S)$ and a target domain $T=(P^T, Y^T)$, we define a Universal Mixing operator $\Psi$ to create an intermediate domain by blending the samples from the two domains:
\begin{equation}
    \Psi(S, T) = (S \odot (\mathbf{1}- \mathcal{M}_S))~\circled{c}~(T\odot \mathcal{M}_T),
\end{equation} 
where $\mathcal{M}_S \in \{\mathcal{M}^{\circledS}_S~|~\circledS=\{\circled{\scriptsize{Sp}},~\circled{\small{In}},~\circled{\footnotesize{Se}}\}\}$ and $\mathcal{M}_T \in \{\mathcal{M}^{\circledS}_T~|~\circledS=\{\circled{\scriptsize{Sp}},~\circled{\small{In}},~\circled{\footnotesize{Se}}\}\}$ represent binary masks indicating where to select from source and target point cloud. $\circled{\scriptsize{Sp}}$, $\circled{\small{In}}$, $\circled{\footnotesize{Se}}$ denote different sampling operators, \ie, spatial mixing, intensity mixing, and semantic mixing. $\mathbf{1}$ is a binary mask filled with ones, $\odot$ denotes Hadamard product, and \circled{c} represents point cloud concatenation. After bidirectionally applying the Universal Mixing operator (denoted as $\mathop{\Psi}\limits ^{\rightleftharpoons}$) to the source data and target data, we can obtain the mixed intermediate domain data and labels as follows:
\begin{equation}
\begin{aligned}
P^{S\rightarrow T} , P^{T\rightarrow S} &= \mathop{\Psi}\limits ^{\rightleftharpoons} (P^S , P^T ),\\
Y^{S\rightarrow T} , Y^{T\rightarrow S} &= \mathop{\Psi}\limits ^{\rightleftharpoons} (Y^S , Y^T ).
\end{aligned}
\end{equation}
\noindent\textbf{Spatial Mixing.} 
For spatial mixing, we adopt a Cylinder coordinate-based partition to both source and target LiDAR point clouds. We first transform the Cartesian coordinates $(x, y, z)$ of the 3D points to the Cylinder coordinate $(\rho, \theta, z)$, where the radius $\rho = \sqrt{(x^2+y^2)}$ is the radial distance from the point to the z-axis and the azimuth $\theta = tan^{-1}(y/x)$ is the angle from x-axis to y-axis indicating the rotation angle in the sensor horizontal plane. Then, the point clouds can be split into a series of non-overlapping spaces along the three dimensions, as shown in Fig.~\ref{fig:mixingfig} (top). For the convenience of exchanging points, we normalize the Cylinder coordinates of two LiDAR scans to obtain the same partition spaces.
The spatial mixing mask can be obtained by dividing the intervals on three coordinate axes: $\mathcal{M}^{\circled{\tiny{Sp}}} = \{(\rho_h<\rho<(\rho_{h}+\Delta\rho))\cap(\theta_i<\theta<(\theta_{i}+\Delta\theta))\cap (z_j<z<(z_j+\Delta z))\}$. Here, $\Delta\rho$, $\Delta\theta$, and $\Delta z$ are preset intervals along three dimensions. $\rho_h$, $\theta_i$, and $z_j$ are randomly picked coordinate values within the corresponding valid ranges. Points covered by the masks will be cut out from the source point cloud and supplemented with the cut from the target points, and vice versa. The labels of the point clouds are sampled and mixed in the same way.

\noindent\textbf{Intensity Mixing.}
For intensity mixing, we conduct the point cloud partition according to the intensity values of 3D points. We first normalize the intensities to the range of $[0,1]$. Then, the intensity mixing mask is obtained by dividing the intensity values into different partitions: $\mathcal{M}^{\circled{\scriptsize{In}}}=\{I_m<I_i<I_m+\Delta I\}$. Here, $\Delta I$ is a preset intensity interval and $I_m$ is a randomly picked intensity value within the valid range. As shown in Fig.~\ref{fig:mixingfig} (middle), the points covered by the masks in both the source and target point clouds (along with their labels) will be cut out and exchanged, alleviating the domain shift caused by intensity variations.

\noindent\textbf{Semantic Mixing.}
Following CosMix \cite{saltori2022cosmix}, we also conduct semantic mixing 
which randomly selects a subset of semantic patches and their labels from the source and target point clouds and exchanges them with each other. Formally, $\mathcal{M}^{\circled{\tiny{Se}}}=\{y_i\in \{\mathit C\}\}$. Here, $y_i$ is the ground truth (or pseudo) semantic label of point $p_i$, and $\mathit C$ is the randomly selected semantic category set. By selecting points from various categories, the mixed point clouds have different semantic distributions with diverse contexts of both domain samples, as shown in Fig. \ref{fig:mixingfig} (bottom).

We employ Universal Mixing in both stages to generate the intermediate domains: $\{P^{S_c\rightarrow B_a}, P^{B_a\rightarrow S_c}\}$ and $\{P^{B_a\rightarrow T_a}, P^{T_a\rightarrow B_a}\}$, as shown in Fig.~\ref{figure:mainfigure}. The intermediate domains comprehensively blend point clouds across domains, which can be utilized to mitigate the domain shift for training domain adaptive and generalizable models.

\begin{algorithm}[!t]
        \caption{The Pipeline of UniMix for DG and UDA.}
   \label{alg:unimix}
\begin{algorithmic}[1]
  \footnotesize
   \STATE {\bfseries Input:} Shuffled labeled source data $(P^{S_c },Y^{S_c} )$, Shuffled unlabeled target data $P^{T_{a}} $, weather simulation generator $\Omega$, Universal Mixing operator $\mathop{\Psi}\limits ^{\rightleftharpoons}$. Student net $\mathit{\Phi^s_1}, \mathit{\Phi^s_2}$ and Teacher net $\mathit{\Phi^t_1}, \mathit{\Phi^t_2}$, and training epochs $E_1, E_2$, for training stage1 and stage2, respectively.\\
   \FOR{$e = 1$ \TO $E_1$}
   \STATE $B_a =(P^{B_a} , Y^{B_a} ) = \Omega(P^{S_c} , Y^{S_c} )$ \COMMENT{\textit{Bridge domain}} \\
   $P^{S_c\rightarrow B_a} , P^{B_a\rightarrow S_c} = \mathop{\Psi}\limits ^{\rightleftharpoons} (P^{S_c} , P^{B_a} )$
   \COMMENT{\textit{Data mixing}} \\
   \STATE $\hat{Y}^{B_a}$ = $\mathit{\Phi^t_1}$($P^{B_a}$) 
   \COMMENT{\textit{Pseudo-label gen.}} \\
   $Y^{S_c\rightarrow B_a} , Y^{B_a\rightarrow S_c} = \mathop{\Psi}\limits ^{\rightleftharpoons} (Y^{S_c} , \hat{Y}^{B_a} )$
   \COMMENT{\textit{Label mixing}} \\
   \STATE $\bar{Y}^{S_c\rightarrow B_a}$ = $\mathit{\Phi^s_1}\big(P^{S_c\rightarrow B_a})$ \COMMENT{\textit{Student training stage1}} \\
    \STATE $\bar{Y}^{B_a\rightarrow S_c}$ = $\mathit{\Phi^s_1}\big(P^{B_a\rightarrow S_c})$ \COMMENT{\textit{Student training stage1}} \\
   \STATE $L^{S_c\rightarrow B_a}$ = $DiceLoss$($\bar{Y}^{S_c\rightarrow B_a}$, $Y^{S_c\rightarrow B_a}$) \COMMENT{\textit{Stage1 loss}}\\
    \STATE $L^{B_a\rightarrow S_c}$ = $DiceLoss$($\bar{Y}^{B_a\rightarrow S_c}$, $Y^{B_a\rightarrow S_c}$) \COMMENT{\textit{Stage1 loss}}\\
    \STATE ${L_{stage1}} = L^{S_c\rightarrow B_a} + L^{B_a\rightarrow S_c}$ \COMMENT{\textit{Stage1 overall loss}}\\
   \STATE Backward($L_{stage1}$), Update($\mathit{\Phi^s_1}$), UpdateEMA($\mathit{\Phi^t_1}$) \\
   \ENDFOR
   \IF{DG}
   \STATE Pred = $\mathit{\Phi^s_1}$($T_a$)  \COMMENT{\textit{Evaluation of DG}}\\
   \ELSIF{UDA}
   
   \FOR{$e = 1$ \TO $E_2$}
   \STATE $P^{B_a\rightarrow T_a} , P^{T_a\rightarrow B_a} = \mathop{\Psi}\limits ^{\rightleftharpoons} (P^{S_{a}} , P^{T_a} )$
   \COMMENT{\textit{Data mixing}} \\
   \STATE $\hat{Y}^{T_a}$ = $\mathit{\Phi^t_2}$($P^{T_a}$) 
   \COMMENT{\textit{Pseudo-label gen.}} \\
   $Y^{B_a\rightarrow T_a} , Y^{T_a\rightarrow B_a} = \mathop{\Psi}\limits ^{\rightleftharpoons} (Y^{S_{a}} , \hat{Y}^{T_a} )$
   \COMMENT{\textit{Label mixing}} \\

   \STATE $\bar{Y}^{B_a\rightarrow T_a}$ = $\mathit{\Phi^s_2}\big(P^{B_a\rightarrow T_a})$ \COMMENT{\textit{Student training stage2}} \\
   \STATE $\bar{Y}^{T_a\rightarrow B_a}$ = $\mathit{\Phi^s_2}\big(P^{T_a\rightarrow B_a})$ \COMMENT{\textit{Student training stage2}} \\
   \STATE $L^{B_a\rightarrow T_a}$ = $DiceLoss$($\bar{Y}^{B_a\rightarrow T_a}$, $Y^{B_a\rightarrow T_a}$) \COMMENT{\textit{Stage2 loss}}\\
    \STATE $L^{T_a\rightarrow B_a}$ = $DiceLoss$($\bar{Y}^{T_a\rightarrow B_a}$, $Y^{T_a\rightarrow B_a}$) \COMMENT{\textit{Stage2 loss}}\\
   \STATE $L_{stage2} = L^{B_a\rightarrow T_a} + L^{T_a\rightarrow B_a}$ \COMMENT{\textit{Stage2 overall loss}}\\
   \STATE Backward($L_{stage2}$), Update($\mathit{\Phi^s_2}$), UpdateEMA($\mathit{\Phi^t_2}$) \\
   \ENDFOR
   \STATE Pred = $\mathit{\Phi^s_2}$($T_a$)  \COMMENT{\textit{Evaluation of UDA}}\\
   \ENDIF
\end{algorithmic}
\end{algorithm}

\subsection{Overall Pipeline}\label{secNets}
The overall architecture is illustrated in Fig.~\ref{figure:mainfigure} and the details are summarized in Algo.~\ref{alg:unimix}. We adopt the teacher-student framework and form a two-stage training pipeline for handling the UDA and DG tasks universally. In each stage, the teacher network is used for producing pseudo-labels for the target data before label mixing and is updated via the exponential moving average (EMA) of the student's weights. 
As shown in Fig.~\ref{figure:mainfigure}, the intermediate domain data $\{P^{S_c\rightarrow B_a}, P^{B_a\rightarrow S_c}\}$ and their labels $\{Y^{S_c\rightarrow B_a} , Y^{B_a\rightarrow S_c}\}$ mixed from the clear-weather source domain $(P^{S_c}, Y^{S_c})$ and the generated Bridge Domain with predicted pseudo-label $(P^{S_{a}}, \hat{Y}^{S_{a}})$ are fed into the student network $\mathit{\Phi^s_1}$ to obtain the predictions $\{\bar{Y}^{S_c\rightarrow B_a}, \bar{Y}^{B_a\rightarrow S_c}\}$. The Dice segmentation loss \cite{jadon2020survey} is adopted to update the learnable weights of the student network $\mathit{\Phi^s_1}$:
\begin{equation}
\begin{aligned}
    L^{S_c\rightarrow B_a} &= DiceLoss(\bar{Y}^{S_c\rightarrow B_a},Y^{S_c\rightarrow B_a}), \\
    L^{B_a\rightarrow S_c} &= DiceLoss(\bar{Y}^{B_a\rightarrow S_c}, Y^{B_a\rightarrow S_c}), \\
    {L_{stage1}} &= L^{S_c\rightarrow B_a} + L^{B_a\rightarrow S_c}.
\end{aligned} 
\end{equation}
Here, $L_{stage1}$ is the overall loss for the first stage of training. Since the data in the adverse weather target domain is inaccessible in the first stage, the student network after the first stage of training can be evaluated on the DG task.

For handling the UDA task, the framework subsequently forwards to the second stage \ie, Bridge-to-Target adaptation, where the Bridge Domain $B_a$ serves the source domain to adapt to the adverse-weather target domain $T_a$. Similar to the first stage, the teacher network is utilized to produce the pseudo-label for the target domain. The mixed intermediate domain data $\{P^{B_a\rightarrow T_a}, P^{T_a\rightarrow B_a}\}$ and labels $\{Y^{B_a\rightarrow T_a} , Y^{T_a\rightarrow B_a}\}$ are obtained through Universal Mixing. The student network $\mathit{\Phi^s_2}$ generates the predictions $\{\bar{Y}^{B_a\rightarrow T_a}, \bar{Y}^{T_a\rightarrow B_a}\}$ and is trained by minimizing the loss ${L_{stage2}} = L^{B_a\rightarrow T_a} + L^{T_a\rightarrow B_a}$. $L_{stage2}$ is similar as $L_{stage1}$, changing the Source-to-Bridge to Bridge-to-Target domain adaptation, as shown in Algo.~\ref{alg:unimix}.
Thus, the student network can be used for evaluation on the UDA task after training. Implementation details can be found in Appendix.

\setlength{\tabcolsep}{2mm}{
\begin{table*}[t]
\centering
\begin{footnotesize}
\resizebox{\textwidth}{!}{%
\begin{tabular}{l|ccccccccccccccccccc|c}
    \toprule
    \rowcolor{black!10} {Method}& \rotatebox{90}{\textbf{car}} & \rotatebox{90}{\textbf{bi.cle}} & \rotatebox{90}{\textbf{mt.cle}} & \rotatebox{90}{\textbf{truck}} & \rotatebox{90}{\textbf{oth-v.}} & \rotatebox{90}{\textbf{pers.}} & \rotatebox{90}{\textbf{bi.clst}} & \rotatebox{90}{\textbf{mt.clst}} & \rotatebox{90}{\textbf{road}} & \rotatebox{90}{\textbf{parki.}} & \rotatebox{90}{\textbf{sidew.}} & \rotatebox{90}{\textbf{oth-g.}} & \rotatebox{90}{\textbf{build.}} & \rotatebox{90}{\textbf{fence}} & \rotatebox{90}{\textbf{veget.}} & \rotatebox{90}{\textbf{trunk}} & \rotatebox{90}{\textbf{terra.}} & \rotatebox{90}{\textbf{pole}} & \rotatebox{90}{\textbf{traf.}} & \textbf{mIoU}\\
    \midrule
   \rowcolor{yellow!20} Oracle & 89.4 & 42.1 & 0.0 & 59.9 & 61.2 & 69.6 & 39.0 & 0.0 & 82.2 & 21.5 & 58.2 & 45.6 & 86.1 & 63.6 & 80.2 & 52.0 & 77.6 & 50.1 & 61.7 & 54.7\\ 
    \midrule
    
  \multicolumn{21}{c}{ SemanticKITTI$\rightarrow$SemanticSTF}\\
  \midrule
     Source-only & 55.9 & 0.0 & 0.2 & 1.9 & 10.9 & 10.3 &  {6.0} & 0.0 & 61.2 & 10.9 & 32.0 & 0.0 & 67.9 & 41.6 & 49.8 & 27.9 & 40.8 &  {29.6} & 17.5  & 24.4 \\
   ADDA~\cite{tzeng2017adda} & 65.6 & 0.0 & 0.0 & 21.0 & 1.3 & 2.8 & 1.3 &  {16.7} & 64.7 & 1.2 & 35.4 & 0.0 & 66.5 & 41.8 &  \textbf{57.2} & 32.6 & 42.2 & 23.3 & 26.4 & 26.3 \\
    Ent-Min~\cite{vu2019advent} & 69.2 & 0.0 & 10.1 & 31.0 & 5.3 & 2.8 &  {2.6} & 0.0 &  {65.9} & 2.6 & 35.7 & 0.0 &  {72.5} &  {42.8} & 52.4 & 32.5 &  \textbf{44.7} & 24.7 & 21.1 & 27.2 \\
    Self-training~\cite{zou2019confidence} &  {71.5} & 0.0 & 10.3 &  \textbf{}{33.1} & 7.4 & 5.9 & 1.3 & 0.0 & 65.1 & 6.5 &  {36.6} & 0.0 & 67.8 & 41.3 & 51.7 &  {32.9} & 42.9 & 25.1 & 25.0 & 27.6  \\
    CoSMix~\cite{saltori2022cosmix} & 65.0 &  \textbf{1.7} &  {22.1} & 25.2 &  {7.7} &  {33.2} & 0.0 & 0.0 & 64.7 &  {11.5} & 31.1 &  {0.9} & 62.5 & 37.8 & 44.6 & 30.5 & 41.1 &  \textbf{30.9} &  {28.6} &  {28.4}  \\
     \textbf{UniMix} &\textbf{75.3 } & 0.9 & \textbf{44.9} &11.7 &\textbf{13.6}  &\textbf{38.2}  & \textbf{50.3} & \textbf{31.9} & \textbf{71.1} & \textbf{15.0} & \textbf{46.4}& \textbf{6.5 }&\textbf{74.3} &\textbf{51.0 } &49.8  &\textbf{36.8}  &34.4 & 25.5 & \textbf{28.9} & \textbf{37.2}\\

    \midrule

  \multicolumn{21}{c}{SynLiDAR$\rightarrow$SemanticSTF}\\
  \midrule
  Source-only& 27.1&  \textbf{3.0}& 0.6& 15.8& 0.1& 25.2& 1.8& 5.6& 23.9& 0.3& 14.6& 0.6& 36.3& 19.9 & 37.9& 17.9& \textbf{41.8}& 9.5& 2.3 & 15.0 \\
    ADDA~\cite{tzeng2017adda} & 55.8 & 0.0 & 3.6 & 26.1 & 1.3 & 25.2 & 7.5 &  9.9 & 17.2 & \textbf{23.4}&4.4 & 0.9 & 43.9 & 18.4 & 45.2 &  21.8 & 33.6 & 28.0 & 19.7  & 20.3 \\
    Ent-Min~\cite{vu2019advent} & 48.3 & 0.1 & 5.6 & \textbf{28.7} & 0.1 & 23.3 &  {2.5} & 19.8 &  19.3 & 6.7 & 22.6 & 1.4 &  46.9 &  20.7 & 43.2 & 25.2 &  34.1 & 26.0 & 22.2 & 20.9 \\
   Self-training~\cite{zou2019confidence} &  50.6 & 0.0 & 6.1 &  {31.0} & 0.5 & 26.0 & 4.8 & 12.0 & 20.7 & 4.6 &  23.5 & 1.5 & 45.3 & 19.5 & 44.6 &  25.0 & 35.1 & \textbf{29.2} & 20.8 & 21.1  \\
    CoSMix~\cite{saltori2022cosmix} & 51.5 &  {0.2} &  {5.0} & 28.1 &  {0.0} &  {26.5} & \textbf{17.0} & 9.9 & 20.2 &  {3.6} &24.6 &  {2.2} &52.6 & 20.6 & 47.5 & 24.3 & 34.6 &  {28.2} &  {24.1} &  22.1 \\
  \textbf{UniMix} & \textbf{73.6} & 0.0 & \textbf{7.9} & 26.9 & \textbf{2.9} & \textbf{29.1} & 13.7 &\textbf{21.8} &\textbf{38.0 } &8.0  & \textbf{26.3} &\textbf{3.4}  & \textbf{56.0} &\textbf{21.2}  &\textbf{56.1} & \textbf{29.6} & 38.0 & 28.2 & \textbf{26.5} &   \textbf{26.7}\\
    \bottomrule
    \end{tabular}}
    \vspace{-7pt}
    \caption{Comparison of SOTA domain adaptation methods on SemanticKITTI$\rightarrow$SemanticSTF and SynLiDAR$\rightarrow$SemanticSTF. SemanticKITTI and SynLiDAR serve as the source domain while SemanticSTF with all four weather conditions serves as the target domain. Our UniMix achieves the best gain of 12.8 and 11.7 mIoU over the source-only model in two benchmarks. 
    }
    \label{tab:uda}
    \end{footnotesize}
    \vspace{-7pt}
\end{table*}
}

\section{Experiments}
We evaluate the adaptive and generalizable ability of UniMix for adverse weather conditions in two tasks, \ie, UDA and DG. For each task, we adopt two benchmark settings, \ie, SemantiKITTI$\rightarrow$SemanticSTF and SynLiDAR$\rightarrow$SemanticSTF, where the clear weather real dataset SemanticKITTI \cite{behley2019semantickitti} and synthetic dataset SynLiDAR \cite{xiao2022transfer} are used as the source domain to adapt or generalize to the adverse-weather target domain SemanticSTF \cite{xiao20233d}, respectively. To further demonstrate the adaptation ability for the synthetic-to-real UDA scenario in clear weather, we also evaluate UniMix in the SynLiDAR$\rightarrow$SemantiKITTI setting. In addition, we employ Universal Mixing for data augmentation to evaluate its impact on increasing the performance of supervised LSS models on the source domain. 

\subsection{Datasets and Metrics}
\noindent \textbf{Datasets.} \textbf{SemanticKITTI} \cite{behley2019semantickitti} is a large-scale LiDAR semantic segmentation dataset, extended from the KITTI Visual Odometry benchmark \cite{geiger2012we}. It covers a wide range of real-world urban scenes captured in German and provides point-wise annotations over 19 semantic categories. We use sequences 00-07 and 09-10 for training and sequence 08 as validation following the official protocol. \textbf{SynLiDAR} \cite{xiao2022transfer} is a large-scale synthetic LiDAR semantic segmentation dataset, containing virtual urban cities, suburban towns, neighborhoods, and harbor scenes. It provides point-wise annotations over 32 semantic categories. Following the command practice \cite{xiao2022transfer,saltori2022cosmix_eccv,saltori2022cosmix,xiao20233d}, we use 19,840 point clouds for training and 1,976 for validation. \textbf{SemanticSTF} \cite{xiao20233d} is an adverse-weather point cloud semantic segmentation dataset that provides point-wise annotations over 21 semantic categories. It contains 2,076 scans from the STF dataset \cite{bijelic2020seeing} that cover various adverse weather conditions including 637 dense fog, 631 light fog, and 114 rain and 694 snow scenes. We follow the official protocol using 1,326 scans for training, and 250 for validation.

\noindent\textbf{Metrics.}
We adopt the evaluation metrics of Intersection over the Union (IoU) for each segmentation class and the mean IoU (mIoU) over all classes. Note that since the label set of SemanticSTF is the same as SemanticKITTI, we map the labels in all the datasets to the 19 common classes of SemanticKITTI following the common practice.

\setlength{\tabcolsep}{1.2mm}{
\begin{table*}[t]
\centering
\begin{footnotesize}
\resizebox{\textwidth}{!}{
\begin{tabular}{l|ccccccccccccccccccc|cccc|c}
 \toprule
   \rowcolor{black!10}Method & \rotatebox{90}{\textbf{car}} & \rotatebox{90}{\textbf{bi.cle}} & \rotatebox{90}{\textbf{mt.cle}} & \rotatebox{90}{\textbf{truck}} & \rotatebox{90}{\textbf{oth-v.}} & \rotatebox{90}{\textbf{pers.}} & \rotatebox{90}{\textbf{bi.clst}} & \rotatebox{90}{\textbf{mt.clst}} & \rotatebox{90}{\textbf{road}} & \rotatebox{90}{\textbf{parki.}} & \rotatebox{90}{\textbf{sidew.}} & \rotatebox{90}{\textbf{oth-g.}} & \rotatebox{90}{\textbf{build.}} & \rotatebox{90}{\textbf{fence}} & \rotatebox{90}{\textbf{veget.}} & \rotatebox{90}{\textbf{trunk}} & \rotatebox{90}{\textbf{terra.}} & \rotatebox{90}{\textbf{pole}} & \rotatebox{90}{\textbf{traf.}} & \rotatebox{90}{\textbf{Dense-fog}} & \rotatebox{90}{\textbf{Light-fog}} &  \rotatebox{90}{\textbf{Rain}} & \rotatebox{90}{\textbf{Snow}} & \textbf{mIoU} \\
  \midrule
   \rowcolor{yellow!20} Oracle & 89.4 & 42.1 & 0.0 & 59.9 & 61.2 & 69.6 & 39.0 & 0.0 & 82.2 & 21.5 & 58.2 & 45.6 & 86.1 & 63.6 & 80.2 & 52.0 & 77.6 & 50.1 & 61.7 & 51.9 & 54.6 & 57.9 & 53.7 & 54.7\\ 
  \midrule
  \multicolumn{25}{c}{SemanticKITTI$\rightarrow$SemanticSTF}\\
  \midrule
  Source-only & 55.9 & 0.0 & 0.2 & 1.9 & 10.9 & 10.3 &  {6.0} & 0.0 & 61.2 & 10.9 & 32.0 & 0.0 & 67.9 & 41.6 & 49.8 & 27.9 & 40.8 &  {29.6} & 17.5 & 29.5 & 26.0 & 28.4 & 21.4 & 24.4 \\
  Dropout~\cite{srivastava2014dropout} &  62.1 & 0.0 &  \textbf{15.5} & 3.0 &  {11.5} & 5.4 & 2.0 & 0.0 & 58.4 & 12.8 & 26.7 & 1.1 & 72.1 & 43.6 & 52.9 &  \textbf{34.2} & 43.5 & 28.4 & 15.5 & 29.3 & 25.6 & 29.4 & 24.8 & 25.7 \\
   Perturbation \cite{xiao20233d}&  {74.4} & 0.0 & 0.0 &  \textbf{23.3} & 0.6 & 19.7 & 0.0 & 0.0 & 60.3 & 10.8 & 33.9 & 0.7 & 72.0 & 45.2 & 58.7 & 17.5 & 42.4 & 22.1 & 9.7 & 26.3 & 27.8 & 30.0 & 24.5 & 25.9 \\
   PolarMix~\cite{xiao2022polarmix} & 57.8 &  {1.8} & 3.8 & 16.7 & 3.7 & 26.5 & 0.0 &  {2.0} & 65.7 & 2.9 & 32.5 & 0.3 & 71.0 &  \textbf{48.7} & 53.8 & 20.5 & 45.4 & 25.9 & 15.8 & 29.7 & 25.0 & 28.6 & 25.6 & 26.0 \\
   MMD~\cite{li2018domain} & 63.6 & 0.0 & 2.6 & 0.1 & 11.4 &  {28.1} & 0.0 & 0.0 &  \textbf{67.0} &  \textbf{14.1} & 37.9 & 0.3 & 67.3 & 41.2 & 57.1 & 27.4 & 47.9 & 28.2 & 16.2 & 30.4 & 28.1 &  {32.8} & 25.2 & 26.9 \\
  PCL~\cite{yao2022pcl} & 65.9 & 0.0 & 0.0 & 17.7 & 0.4 & 8.4 & 0.0 & 0.0 & 59.6 & 12.0 & 35.0 &  \textbf{1.6} &  \textbf{74.0} & 47.5 &  \textbf{60.7} & 15.8 &  \textbf{48.9} & 26.1 &  {27.5} & 28.9 & 27.6 & 30.1 & 24.6 & 26.4 \\
 PointDR~\cite{xiao20233d} &  67.3 & 0.0 & 4.5 & 19.6 & 9.0 & 18.8 & 2.7 & 0.0 & 62.6 & 12.9 &  \textbf{38.1} & 0.6 & 73.3 & 43.8 & 56.4 & 32.2 & 45.7  & 28.7 & 27.4 &  {31.3} &  {29.7} & 31.9 &  {26.2} &  {28.6} \\
 \textbf{UniMix} & \textbf{82.7 } & \textbf{6.6 }& 8.6 & 4.5 &  \textbf{15.1}& \textbf{35.5} & \textbf{15.5} & \textbf{37.7 }& 55.8 &  10.2 & 36.2& 1.3 & 72.8 & 40.1 & 49.1 & 33.4  &34.9  &\textbf{23.5}  & \textbf{33.5}  & \textbf{34.8}  & \textbf{30.2} &\textbf{34.9} & \textbf{30.9} & \textbf{31.4}  \\
  \midrule
  \multicolumn{25}{c}{SynLiDAR$\rightarrow$SemanticSTF}\\
  \midrule
 Source-only& 27.1&  \textbf{3.0}& 0.6& 15.8& 0.1& 25.2& 1.8& 5.6& 23.9& 0.3& 14.6& 0.6& 36.3& 19.9 & 37.9& 17.9& 41.8& 9.5& 2.3 & 16.9 & 17.2 & 17.2 & 11.9 & 15.0 \\
   Dropout~\cite{srivastava2014dropout} & 28.0 &  \textbf{3.0} & 1.4 & 9.6 & 0.0 & 17.1 & 0.8 & 0.7 &  {34.2} & 6.8 & 19.1 & 0.1 & 35.5 & 19.1 & 42.3 & 17.6 & 36.0 & 14.0 & 2.8 & 15.3 & 16.6 & 20.4 & 14.0 & 15.2 \\
 Perturbation \cite{xiao20233d}& 27.1 & 2.3 & 2.3 & 16.0 & 0.1 & 23.7 & 1.2 & 4.0 & 27.0 & 3.6 & 16.2 & 0.8 & 29.2 & 16.7 & 35.3 & \textbf{22.7} & 38.3 &  {17.9} & 5.1 & 16.3 & 16.7 & 19.3 & 13.4 & 15.2 \\
  PolarMix~\cite{xiao2022polarmix} &  {39.2} & 1.1 & 1.2 & 8.3 &  {1.5} & 17.8 & 0.8 & 0.7 & 23.3 & 1.3 & 17.5 & 0.4 & 45.2 & 24.8 &  {46.2} & 20.1 & 38.7 & 7.6 & 1.9 & 16.1 & 15.5 & 19.2 & 15.6 & 15.7 \\
  MMD~\cite{li2018domain} &  25.5 & 2.3 & 2.1 & 13.2 & 0.7 & 22.1 & 1.4 & 7.5 & 30.8 & 0.4 & 17.6 & 0.2 & 30.9 & 19.7 & 37.6 & 19.3 &  \textbf{43.5} & 9.9 & 2.6 & 17.3 & 16.3 & 20.0 & 12.7  & 15.1 \\
 PCL~\cite{yao2022pcl} &  30.9 &  0.8 &  1.4 &  10.0 &  0.4 &  23.3 &   \textbf{4.0} &   {7.9} &  28.5 &  1.3 &   {17.7} &   {1.2} &  39.4 &  18.5 &  40.0 &  16.0 &  38.6 &  12.1 &  2.3 & 17.8 & 16.7 & 19.3 & 14.1 & 15.5 \\
   PointDR \cite{xiao20233d} & 37.8 & 2.5 &  {2.4} &  \textbf{23.6} & 0.1 &  {26.3} & 2.2 & 3.3 & 27.9 &  \textbf{7.7} & 17.5 & 0.5 &  {47.6} &  \textbf{}{25.3} & 45.7 &  {21.0} & 37.5 &  {17.9} &  {5.5} &  {19.5} &  {19.9} &  {21.1} &  {16.9} &  {18.5} \\
\textbf{UniMix} &\textbf{65.4} & 0.1&\textbf{ 3.9} & 16.9 & \textbf{5.3} & \textbf{32.3} & 2.0 & \textbf{19.3} &\textbf{ 52.1} &  5.0 & \textbf{27.3} &\textbf{3.0}  &  \textbf{49.4} &20.3  &\textbf{58.5} & \textbf{22.7} & 23.2 & \textbf{26.9}  & \textbf{10.4}  &\textbf{24.3}   &\textbf{22.9}   & \textbf{26.1} &  \textbf{20.9 }& \textbf{23.4} \\
    \bottomrule
    \end{tabular}}
    \vspace{-7pt}
    \caption{Comparison of SOTA domain generalization methods on SemanticKITTI$\rightarrow$SemanticSTF and SynLiDAR$\rightarrow$SemanticSTF.}
    \label{tab:dg_stf}
  \end{footnotesize}
  \vspace{-0.3cm}
\end{table*}
}

\begin{table}[t]
    \centering
    \begin{footnotesize}
    \resizebox{0.48\textwidth}{!}{
    \begin{tabular}{c|ccccc|ccccc}
    \toprule
Bridge &\multicolumn{5}{c|}{\cellcolor{red!10} UDA mIoU}&\multicolumn{5}{c}{\cellcolor{blue!10}DG mIoU}\\
         Domain& \cellcolor{red!10}D-fog &\cellcolor{red!10}L-fog &\cellcolor{red!10}Rain &\cellcolor{red!10}Snow &\cellcolor{red!10}All& \cellcolor{blue!10}D-fog &\cellcolor{blue!10}L-fog &\cellcolor{blue!10}Rain &\cellcolor{blue!10}Snow &\cellcolor{blue!10}All\\
    \midrule
    None &39.8  &31.6 &35.3&29.7 &  31.6&29.5 & 26.0 & 28.4 & 21.4 & 24.4 \\
        L-fog &  38.2& 32.0&37.9& 31.0&34.3 &32.6  &28.1 &29.6 &22.1 &27.1\\
        D-fog &  40.0& 33.0& 38.4&31.3 &34.7&33.7  &26.4 &29.4 &22.5 &27.5\\
        Rain&  40.2& 31.9& 39.3& 31.5&34.9 & 34.5 &25.9 &32.9 &23.2 & 28.1\\
    Snow & 39.5 &32.5 & 37.3&31.6 &35.5 & 31.8 &28.7 &32.9 &24.5 &28.4 \\
     
     Mixed-fog &40.3  & 33.3&38.7 & 32.2& 36.3& 35.7 & 26.8& 32.5& 23.5&29.7\\
    \midrule
    All &40.2& 34.0 &37.5 & 33.2& 37.2&34.8  & 30.2 &34.9 &30.9&31.4 \\
    
    \bottomrule
    \end{tabular}}
    \vspace{-7pt}
    \caption{Ablation study of Bridge Domain compositions used in UniMix for UDA and DG on SemanticKITTI$\rightarrow$SemanticSTF. ``D-fog'' and ``L-fog'' denote dense fog and light fog, respectively.}
    \label{tab:ablation_bridgedomain}
    \end{footnotesize}
    \vspace{-7pt}
\end{table}

\subsection{UniMix for Unsupervised Domain Adaptation}
\noindent\textbf{Setup.} 
In the UDA task, we evaluate UniMix on two benchmarks: SemantiKITTI$\rightarrow$SemanticSTF and SynLiDAR$\rightarrow$ SemanticSTF. 
We compare UniMix with four representative UDA approaches including ADDA~\cite{tzeng2017adda}, entropy minimization (Ent-Min)~\cite{vu2019advent}, Self-training~\cite{zou2019confidence}, and CoSMix~\cite{saltori2022cosmix}. For a fair comparison, all these methods are trained using the MinkowskiNet \cite{choy20194d} as the 3D segmentation backbone. 

\noindent\textbf{Performance.} As shown in Table \ref{tab:uda}, we take the whole SemanticSTF dataset (including all adverse weather scenes) as the target domain to show the performance of UDA methods on the validation set, adapting the model from the SemanticKITTI source (top part) and SynLiDAR source (bottom part), respectively. The results of the \textit{Oracle} model and the \textit{Source-only} model are the upper bound and lower bound for each adaptation scenario. All UDA methods achieve superior performance than the source-only model in both benchmarks. Our UniMix achieves the best gain of 12.8 mIoU for SemantiKITTI$\rightarrow$SemanticSTF adaptation and 11.7 mIoU for SynLiDAR$\rightarrow$SemanticSTF adaption, with an improvement of 8.8 mIoU and 4.6 mIoU over the SOTA synthetic-to-real adaptation method CoSMix \cite{saltori2022cosmix}. Additionally, the results indicate that the adaptation from a synthetic dataset is more difficult for all UDA methods than that from a real dataset, which may be attributed to the additional domain shift between synthetic and real-world data.



\subsection{UniMix for Domain Generalization}
\noindent\textbf{Setup.} 
Different from the UDA setting, the target domain is supposed to be inaccessible during training for DG. 
Following PointDR \cite{xiao20233d}, we compare UniMix with three augmentation-based methods (\ie, Dropout \cite{srivastava2014dropout}, Noise perturbation, PolarMix \cite{xiao2022polarmix}) and two representative 2D DG methods (\ie, MMD \cite{li2018domain} and PCL \cite{yao2022pcl}). We evaluate the methods on two benchmarks: SemantiKITTI $\rightarrow$SemanticSTF and SynLiDAR$\rightarrow$SemanticSTF. 

\noindent\textbf{Performance.} In Table \ref{tab:dg_stf}, we present the generalization results of different methods on the two benchmarks.
Source-only model is trained on the source SemanticKITTI or SynLiDAR dataset without applying any generalization techniques. It performs very poorly in the adverse-weather target domain, due to the large domain gap. The augmentation-based methods and 2D DG methods improve the performance over the Source-only baseline but the gains are very limited in the range of 1.3-2.5 mIoU for SemanticKITTI$\rightarrow$SemanticSTF generalization and 0.1-0.7 mIoU for SynLiDAR$\rightarrow$SemanticSTF generalization. The state-of-the-art method PointDR achieves a gain of 4.2 mIoU and 3.5 mIoU but still lags behind our UniMix, which achieves a gain of 7.0 and 8.4 mIoU over the Source-only baseline. We also show the generalization results for each weather in the right part of Table \ref{tab:dg_stf}, where UniMix also achieves the best performance among all methods.

\subsection{Ablation Study}
We conduct ablation studies of UniMix on SemanticKITTI $\rightarrow$SemanticSTF for both UDA and DG tasks, focusing on three key components of UniMix: the Bridge Domain, Universal Mixing, and the two-stage training architecture. 

\begin{table}[t]
    \centering
    \begin{footnotesize}
    \resizebox{0.48\textwidth}{!}{
    \begin{tabular}{c|ccccc|ccccc}
    \toprule
    Mixing &\multicolumn{5}{c|}{\cellcolor{red!10}UDA mIoU}&\multicolumn{5}{c}{\cellcolor{blue!10}DG mIoU}\\
         Method& \cellcolor{red!10}D-fog &\cellcolor{red!10}L-fog &\cellcolor{red!10}Rain &\cellcolor{red!10}Snow &\cellcolor{red!10}All& \cellcolor{blue!10}D-fog &\cellcolor{blue!10}L-fog &\cellcolor{blue!10}Rain &\cellcolor{blue!10}Snow &\cellcolor{blue!10}All\\
    \midrule
    Source-only&26.9& 25.2&27.7& 23.5& 24.3&29.5& 26.0&28.4& 21.4& 24.4\\
    \midrule
        Spatial  &41.6  &34.6 &35.7&27.8 & 35.2 & 33.0 &28.3 &27.9 &23.3&26.4\\
       Intensity  &41.2  &34.5 &37.4&28.8 &34.9 &33.8  & 25.6&30.0& 18.9&25.8  \\
        Semantic  &41.7  & 34.3 &29.6&36.3 &35.4 & 37.4 &26.7& 27.8 & 22.3& 26.7\\
    \midrule
    All &40.2& 34.0 &37.5 & 33.2&37.2 &34.8  & 30.2 &34.9 &30.9&31.4\\
    \bottomrule
    \end{tabular}}
    \vspace{-7pt}
    \caption{Ablation study of mixing operators for UDA and DG on SemanticKITTI$\rightarrow$SemanticSTF.}
    \label{tab:ablationmixing}
    \end{footnotesize}
    \vspace{-0.3cm}
\end{table}

\setlength{\tabcolsep}{5.2mm}{
\begin{table}[t]
    \centering
    \begin{scriptsize}
    \begin{tabular}{c|c|c}
    \toprule
         Method& \cellcolor{red!10}UDA mIoU &\cellcolor{blue!10}DG mIoU\\
    \midrule
    Source-only&24.4&24.4\\
    \midrule
    
    CosMix \cite{saltori2022cosmix} &35.4&26.7\\
    PolarMix \cite{xiao2022polarmix} &33.0&26.0\\
    LaserMix \cite{kong2023lasermix} &33.3&26.9\\
    Universal Mixing &\textbf{37.2}&\textbf{31.4}\\
    \bottomrule
    \end{tabular}
    \vspace{-7pt}
    \caption{Comparison of mixing methods in our UniMix pipeline for UDA and DG on SemanticKITTI$\rightarrow$SemanticSTF.}
    \label{tab:ablation_replacemixing}
    \end{scriptsize}
    \vspace{-10pt}
\end{table}
}
\setlength{\tabcolsep}{2.2mm}{
\begin{table*}[t]
    \centering
    \begin{footnotesize}
    \resizebox{\textwidth}{!}{%
    \begin{tabular}{l|ccccccccccccccccccc|c}
        \toprule
        \rowcolor{black!10}{Method} & \rotatebox{90}{ \textbf{car}} & \rotatebox{90}{ \textbf{bi.cle}} & \rotatebox{90}{ \textbf{mt.cle}} & \rotatebox{90}{ \textbf{truck}} & \rotatebox{90}{ \textbf{oth-v.}} & \rotatebox{90}{ \textbf{pers.}} & \rotatebox{90}{ \textbf{bi.clst}} & \rotatebox{90}{ \textbf{mt.clst}} & \rotatebox{90}{ \textbf{road}} & \rotatebox{90}{ \textbf{park.}} & \rotatebox{90}{ \textbf{sidew.}} & \rotatebox{90}{ \textbf{oth-g.}} & \rotatebox{90}{ \textbf{build.}} & \rotatebox{90}{ \textbf{fence}} & \rotatebox{90}{ \textbf{veget.}} & \rotatebox{90}{ \textbf{trunk}} & \rotatebox{90}{ \textbf{terra.}} & \rotatebox{90}{ \textbf{pole}} & \rotatebox{90}{ \textbf{traf.}} &  \textbf{mIoU} \\
        \midrule
        
       \rowcolor{yellow!20} Source-only &   42.0 &  5.0 &  4.8 &  0.4 &  2.5 &  12.4 &  43.3 &  1.8 &  48.7 &  4.5 &  31.0 &  0.0 &  18.6 &  11.5 &  60.2 &  30.0 &  48.3 &  19.3 &  3.0 &  20.4 \\
        \midrule

        ADDA~\cite{tzeng2017adda} & 52.5 & 4.5 & 11.9 & 0.3 & 3.9 & 9.4 & 27.9 & 0.5 & 52.8 & 4.9 & 27.4 & 0.0 & 61.0 & 17.0 & 57.4 & 34.5 & 42.9 & 23.2 & 4.5 & 23.0\\
        Ent-Min~\cite{vu2019advent} & 58.3 & 5.1 & 14.3 & 0.3 & 1.8 & 14.3 &  \textbf{44.5} & 0.5 & 50.4 & 4.3 & 34.8 & 0.0 & 48.3 & 19.7 & 67.5 & 34.8 &  \textbf{52.0} & 33.0 & 6.1 & 25.8 \\
        ST~\cite{zou2019confidence} & 62.0 & 5.0 & 12.4 & 1.3 & 9.2 & 16.7 & 44.2 & 0.4 & 53.0 & 2.5 & 28.4 & 0.0 & 57.1 & 18.7 & 69.8 &  {35.0} & 48.7 & 32.5 & 6.9 & 26.5 \\
        PCT~\cite{xiao2022transfer} & 53.4 & 5.4 & 7.4 & 0.8 & 10.9 & 12.0 & 43.2 & 0.3 & 50.8 & 3.7 & 29.4 & 0.0 & 48.0 & 10.4 & 68.2 & 33.1 & 40.0 & 29.5 & 6.9 & 23.9 \\
        ST-PCT~\cite{xiao2022transfer} & 70.8 &  {7.3} & 13.1 & 1.9 & 8.4 & 12.6 & 44.0 & 0.6 & 56.4 & 4.5 & 31.8 & 0.0 &  {66.7} &  \textbf{23.7} & \textbf{73.3} & 34.6 & 48.4 &  {39.4} & 11.7 & 28.9 \\
        CosMix \cite{saltori2022cosmix} & {75.1} & 6.8 & {29.4} & {27.1} & \textbf{11.1} & {22.1} & 25.0 & \textbf{24.7} & \textbf{79.3} & {14.9} & \textbf{46.7} & \textbf{0.1} & 53.4 & 13.0 & 67.7 & 31.4 & 32.1 & 37.9 &{13.4} & 32.2\\
        PolarMix \cite{xiao2022polarmix} &76.3  &\textbf{8.4}  &17.8  &3.9  &6.0 &\textbf{26.6}  &40.8  & 15.9  &70.3   &  0.0 &  44.4 & 0.0  & \textbf{68.4}  &  14.7 & 69.6  & \textbf{38.1}  &  37.1 & \textbf{40.6}  & 10.6  &31.0\\
           \midrule
           UniMix&\textbf{80.3}  &6.1  & \textbf{32.5} &\textbf{29.2}  &10.6 &23.7  &30.0  & 24.5  &62.2   &\textbf{15.9}   & 46.5  &\textbf{0.1}   & 60.9  & 15.8  &70.7   & 34.9  & 41.2  & 38.6  &\textbf{17.1 }  &\textbf{33.7} \\
        \bottomrule

    \end{tabular}
    }
    \vspace{-7pt}
    \caption{Comparison of SOTA domain adaptation methods on SynLiDAR$\rightarrow$SemanticKITTI adaptation scenario. The superior performance demonstrates the potential of UniMix to perform as a general UDA method for LSS.}
    \vspace{-0.2cm}
    \label{tab:syn2real}
    \end{footnotesize}
\end{table*}}
\setlength{\tabcolsep}{2mm}{
\begin{table*}[t]
    \centering
    \begin{footnotesize}
    \resizebox{\textwidth}{!}{%
    \begin{tabular}{l|l|ccccccccccccccccccc|l}
        \toprule
        \rowcolor{black!10}{Dataset}&Method  & \rotatebox{90}{\textbf{car}} & \rotatebox{90}{\textbf{bi.cle}} & \rotatebox{90}{\textbf{mt.cle}} & \rotatebox{90}{\textbf{truck}} & \rotatebox{90}{\textbf{oth-v.}} & \rotatebox{90}{\textbf{pers.}} & \rotatebox{90}{\textbf{bi.clst}} & \rotatebox{90}{\textbf{mt.clst}} & \rotatebox{90}{\textbf{road}} & \rotatebox{90}{\textbf{parki.}} & \rotatebox{90}{\textbf{sidew.}} & \rotatebox{90}{\textbf{oth-g.}} & \rotatebox{90}{\textbf{build.}} & \rotatebox{90}{\textbf{fence}} & \rotatebox{90}{\textbf{veget.}} & \rotatebox{90}{\textbf{trunk}} & \rotatebox{90}{\textbf{terra.}} & \rotatebox{90}{\textbf{pole}} & \rotatebox{90}{\textbf{traf.}} & \textbf{mIoU}\\
        \midrule
        
       \multirow{2}{*}{SemanticKITTI}& \cellcolor{yellow!20}MinkNet \cite{choy20194d}  &\cellcolor{yellow!20}95.7 &\cellcolor{yellow!20} 3.7 &\cellcolor{yellow!20} 44.9 &\cellcolor{yellow!20} 53.2 &\cellcolor{yellow!20} 42.1 &\cellcolor{yellow!20} 53.7 &\cellcolor{yellow!20} 68.9 &\cellcolor{yellow!20} 0.0 &\cellcolor{yellow!20} 92.8 &\cellcolor{yellow!20} 43.0 &\cellcolor{yellow!20} 80.0 &\cellcolor{yellow!20} 1.8 &\cellcolor{yellow!20} 90.5 &\cellcolor{yellow!20} 60.0 &\cellcolor{yellow!20} 87.4 &\cellcolor{yellow!20} 64.5 &\cellcolor{yellow!20} 73.3 &\cellcolor{yellow!20} 62.1 &\cellcolor{yellow!20} 43.7 &\cellcolor{yellow!20} 55.9\\
        &+UM &90.2 & 50.5 & 74.4 &82.3 & 70.2 & 62.7    & 67.5 & 0.1 &73.1   & 56.9  &68.5 &21.4  &86.5 &43.2  & 84.6 &  70.1& 51.5 &  71.3 & 43.5  & $61.5_{+5.6}$\\
        \midrule
        \multirow{2}{*}{SynLiDAR}& \cellcolor{yellow!20}MinkNet \cite{choy20194d} &\cellcolor{yellow!20} 77.5 &\cellcolor{yellow!20} 12.0 &\cellcolor{yellow!20} 83.6 &\cellcolor{yellow!20} 87.5 &\cellcolor{yellow!20} 88.6 &\cellcolor{yellow!20} 70.6 &\cellcolor{yellow!20} 42.7 &\cellcolor{yellow!20} 94.1 &\cellcolor{yellow!20} 80.7 &\cellcolor{yellow!20} 76.7 &\cellcolor{yellow!20} 69.5 &
        \cellcolor{yellow!20}93.7 &\cellcolor{yellow!20} 92.6&\cellcolor{yellow!20} 80.3 &\cellcolor{yellow!20} 85.2 &\cellcolor{yellow!20}86.1 &\cellcolor{yellow!20} 65.4 &\cellcolor{yellow!20} 69.2 &\cellcolor{yellow!20} 75.5 &\cellcolor{yellow!20} 75.3\\ 
        &+UM &75.7  &70.7  & 80.8 & 86.4 &86.9  &  65.7& 89.1 & 94.4&  93.0& 73.1 & 80.8 & 92.6 & 93.8 & 77.0 &87.8 & 83.1 & 74.2 & 65.4 & 63.7 &  $80.7_{+5.4}$\\
        \bottomrule

    \end{tabular}
    }
    \vspace{-5pt}
    \caption{Universal Mixing (denoted as ``UM'') can also be used as a data augmentation technique to improve the performance of supervised semantic segmentation. 
    }
    \label{tab:dataAug}
    \end{footnotesize}
    \vspace{-0.3cm}
\end{table*}
}
\noindent\textbf{Bridge Domain Compositions.} Our method produces four frequent weather conditions in real life, \ie, dense fog, light fog, rain, and snow to form the Bridge Domain. As shown in Table \ref{tab:ablation_bridgedomain}, we ablate the impact of the simulated weather conditions in generating the Bridge Domain, including four single-weather settings and a mixed fog setting, where 50\% dense fog data and 50\% light fog data are generated. We show the results evaluated on the whole validation set with all four weather conditions as well as on the subsets of each weather condition. As can be seen, UniMix with any of the choices for generating the Bridge Domain achieves a gain of 2.7-5.6 mIoU (2.7-7.0 mIoU) for UDA (DG) over the vanilla UniMix baseline without using the Bridge Domain. Among all the choices, the mixed-fog brings the largest improvement in both tasks, while the light-fog produces the smallest improvement. This can be attributed to the small perturbations brought by the light-fog simulation, while the mixed-fog domain integrates diverse disturbance patterns via both dense-fog and light-fog simulation. We also observe a relatively high performance when the simulated weather in the Bridge Domain is the same as the weather of the validation subset, which is reasonable and implies the necessity of simulating all weather conditions in the Bridge Domain as validated by the last row.

\noindent\textbf{Sample Mixing Methods.}
Universal Mixing contains three kinds of mixing methods, \ie, spatial, intensity, and semantic mixing. As shown in Table \ref{tab:ablationmixing}, we ablate the choice of mixing methods. According to the results, each of the three mixing methods enhances the performance of the Source-only baseline on both UDA and DG tasks. Semantic mixing delivers the largest improvement on the whole validation set while performing worse in the rain subset than other mixing methods. Only using intensity mixing brings the smallest improvement while performing the best in the rain subset. We attribute this to the distinct reflectance attenuation caused by different weather conditions. By incorporating all the mixing methods together, UniMix achieves the best performance on both tasks. 

In addition, we compare our Universal Mixing with some representative point cloud mixing methods, including LaserMix \cite{kong2023lasermix} proposed for semi-supervised LSS, PolarMix \cite{xiao2022polarmix} for point cloud data augmentation, and CosMix \cite{saltori2022cosmix} for synthetic-to-real domain adaptation. We replace our Universal Mixing with these alternatives in our UniMix pipeline and evaluate them on both UDA and DG tasks. As shown in Table \ref{tab:ablation_replacemixing}, the proposed Universal Mixing achieves the best performance, showing its superiority over others.

\noindent\textbf{Training Stage.}
Multi-stage training is widely used in 2D UDA semantic segmentation methods \cite{zhang2019category, Zhang_2021_CVPR}. Our method adopts a two-stage training framework as well. We show the UDA results without the Bridge Domain and the first-stage training, denoted as ``None" in the first row in Table \ref{tab:ablation_bridgedomain}. The performance is decreased but still superior to other SOTA methods. 
In the DG setting, this version is the Source-only model, without using any Bridge or target domain data. It is also noteworthy that our two-stage UniMix for UDA can generate a DG model after the first training stage, providing a versatile solution and holding practical significance.

\subsection{UniMix in Clear Weather}
Although UniMix is tailored for clear-to-adverse weather domain adaptation and generalization, we also examine UniMix in synthetic-to-real adaptation in clear weather. In Table \ref{tab:syn2real}, UniMix achieves 33.7 mIoU with a gain of 13.3 mIoU over the Source-only model. Its superior performance over the SOTA methods validates the effectiveness of UniMix for synthetic-to-real adaptation, implying that it could be used as a general method for LSS UDA tasks.

\subsection{Universal Mixing as Data Augmentation}

We also use Universal Mixing as a data augmentation technique in supervised LSS tasks, using the widely adopted MinkowskiNet \cite{choy20194d} as our baseline. Results in Table \ref{tab:dataAug} show significant performance enhancement with Universal Mixing on both datasets compared to the baseline model.
\section{Conclusion}

In this paper, we introduce UniMix, a universal method designed to enhance the adaptive and generalizable capabilities of LiDAR semantic segmentation models in adverse weather conditions. First, a Bridge Domain is constructed, capturing scene characteristics of the clear-weather source domain with adverse weather conditions to learn weather-robust representations. Universal Mixing is then proposed and employed in two stages to blend samples from the source and target domains, considering spatial, intensity, and semantic distribution, to learn domain invariant features. UniMix proves highly effective in UDA and DG tasks, significantly outperforming state-of-the-art methods.
\newpage
{
    \small
    \bibliographystyle{ieeenat_fullname}
    \bibliography{main}
}


\end{document}